# Alignment of Microtubule Imagery

Feiyang Yu, Ard Oerlemans, and Erwin M. Bakker
Leiden University, The Netherlands
{fionayu, aoerlema, erwin}@liacs.nl

**ABSTRACT**

*This work discusses preliminary work aimed at simulating and visualizing the growth process of a tiny structure inside the cell---the microtubule. Difficulty of recording the process lies in the fact that the tissue preparation method for electronic microscopes is highly destructive to live cells. Here in this paper, our approach is to take pictures of microtubules at different time slots and then appropriately combine these images into a coherent video. Experimental results are given on real data.*

## 1. INTRODUCTION

Image registration is a crucial process for understanding of the semantic content of domain specific images. Information gained from two or more images acquired in the track of a specific event is usually of a complementary nature. Therefore proper integration of that information provides a sound basis for later analysis by experts.

Microtubules are tiny structures (24nm diameter) found in cells [1,2]. Understanding of the development path of that structure still remains an open issue in the bio-informatics research community. With the help of electronic microscopy (EM) technology, taking a single image of a microtubule at a specific time slot is possible. Registering and combining individual images from microtubules according to their temporal order can potentially illuminate our understanding of their patterns of growth. The challenge lies in the fact that tissue preparation method used by EM is highly destructive for living cells. Therefore, the microscopy images were very noisy due to the short amount of time the radiation can be sent through the cell.

The remainder of the paper is organized as follows. In Section 2, a brief review of related work in literature is presented. Section 3 elaborates our method for registering microtubule images. The registered images later are combined to produce a proper video. Experimental results are presented in Section 4. Finally, we conclude this paper in Section 5.

## 2. RELATED WORK

Extensive studies on image registration have been presented in the literature. A comprehensive survey of image registration methods was published in 1992 by Brown [6]. More recent developments of image registration are reviewed by Zitova etc al in [7].

Registration methods can be categorized according to their selection of feature detection methods [7]. The first category is the so-called feature-based methods. This kind of method extracts salient features, e.g., significant regions, lines or points, from sensed and reference images, and aligns two images by the overlap criterion of the selected features. On the other hand, the second category focuses on feature matching rather than on feature detection. Windows of predefined size or even entire images are used for the correspondence estimation during the registration step.

Due to the highly noisy characteristics of our microtubule images, we also follow the area-based image registration method. The typical representative of the area-based method is the normalized cross-correlation (NCC) method and its modifications [3, 4, 8]. This similarity measure is ideal for matching images that differ by a translation of the intensity map. A generalized version of NCC for geometrically more deformed images was also presented in literature. Hanaizumi et al. proposed to compare the NCC criterion for each assumed geometric transformation of the sensed image window [9]. Psarakis et al. interpolated the candidate windows of the matching image and used the classical zero mean NCC function to measure stereo correspondence [3, 4, 10]. Although the NCC similarity measure is rather easy to implement, the computation load grows very fast with the complexity of transformations

## 3. OUR METHOD

Because of the great variety of visual appearances of images used in this work, appropriate image registration is crucial for the success of our simulation. Typical images of development stages of microtubules were taken at different times, rotations, translations, and viewpoints. How to align those images properly directly determines the usefulness of our video.

One important feature of microtubule structure is their polarity. Ends of a microtubule are designated as plus and minus respectively. The plus-end plays an import role in growth of microtubules, and is the dominant object of our images. For easy registration, manual preprocessing of images is conducted to crop all images into equivalent size and with the plus-end in the center of an image.

After the manual manipulation and enhancement of the contrast of images, we designate an appropriate image as the initial frame of the output video. Candidate images are then normalized to get zero-mean and unit intensity. The initial frame also functions as the reference frame for image registration.

The fundamental problem with image registration is how to find a type of spatial transformation to properly overlay two images. Although many types of variations may be present in each image, the registration technique must select the class of transformation, which will remove only the spatial distortions between images due to difference in acquisition conditions. Our assumption is that candidate images can be mapped to the reference frame by a rotation transformation. That is, we assume the major object, the plus-end, retains its relative shape and size across samples and acquisition conditions. In homogeneous coordinates, rotation of an image can be characterized by:

$$\begin{pmatrix} x' \\ y' \\ 1 \end{pmatrix} = \begin{pmatrix} \cos\theta & -\sin\theta & 0 \\ \sin\theta & \cos\theta & 0 \\ 0 & 0 & 1 \end{pmatrix} \begin{pmatrix} x \\ y \\ 1 \end{pmatrix} \quad (1)$$

where $(x, y)^T$ denotes the coordinates of the original point and $(x', y')^T$ denotes the coordinates of the transformed point. $\theta$ is the rotation angle.

To find the optimal rotation angle, we employ normalized cross correlation as the similarity measure. Normalized cross correlation is the most extensively used similarity measure for template matching. Images are naturally digitized in the Cartesian coordinates. For easy manipulation of image rotation, every image is converted to the polar coordinates. The polar mapping of an image is illustrated in the following figure.

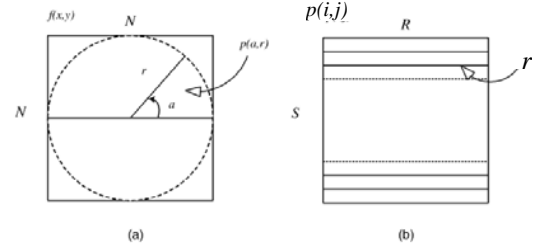

**Figure 1** Log-polar transform of $N*N$ image (f(x,y)) into $S*R$ polar image (p(i,j)) by first (a) using radius as scanline for sampling N times the circle to produce a polar form p(a,r), and (b) applying quantization on all radii to produce the polar image.

In the polar coordinates, NCC between two images $I_1$ and $I_2$ can be evaluated as the following:

$$NCC_{I_1,I_2} = \frac{\sum_r \sum_\theta \left(I_1(r,\theta)-\bar{I}_1(r,\theta)\right)\left(I_2(r,\theta)-\bar{I}_2(r,\theta)\right)}{\sqrt{\sum_r \sum_\theta \left(I_1(r,\theta)-\bar{I}_1(r,\theta)\right)^2 \sum_r \sum_\theta \left(I_1(r,\theta)-\bar{I}_1(r,\theta)\right)^2}} \quad (2)$$

where $\bar{I}_1$ and $\bar{I}_2$ are the mean intensities of images $I_1$ and $I_2$ respectively. NCC takes on values in $[-1,1]$ and is invariant to brightness shifts and contrast scaling. By shifting the polar version of the a candidate image $I_{candidate}$ horizontally pixel by pixel, the NCC value of the image to $I_{reference}$ will arrive a maximal value, which correspondes to the optimal rotation angle. The angle can be formalized as:

$$\hat{\theta} = \arg\max_\theta (NCC_{I_{reference} I'_{candidate}}) \quad (3)$$

where $I'_{candidate}$ denotes the shifted version of $I_{candidate}$.

The last step for the creation of a video is to find pairwise highly correlated images. We calculate the NCC value again, but now for only the center part of the normalized image. The center of the image contains the plus-end of the microtubule. Our assumption is that images with high correlation values represent almost the same stage in the growth process of a microtubule. The correlation value is scaled to a probability value, and employed to as a measure to select susccessive images. An illustraion of this procedure is given in the following table.

Table **1** Illustration for integrating images into a video

| Probability | $I_1$ | $I_2$ | $I_3$ | $I_4$ |
|---|---|---|---|---|
| $I_1$ | 1 | 0.8 | 0.4 | 0.6 |
| $I_2$ | 0.8 | 1 | 0.5 | 0.7 |
| $I_3$ | 0.4 | 0.5 | 1 | 0.6 |
| $I_4$ | 0.6 | 0.7 | 0.6 | 1 |

Suppose image $I_1$ is taken as the starting frame for the video file. The next image can be one out of $I_2$, $I_3$ and $I_4$. Looking up the probabilities Table **1**, it is easy to see that images $I_2$ ($P(I_2|I_1)=0.8$) and $I_4$ ($P(I_2|I_4)=0.6$) have the highest probability of being selected as the next image. By repeating this process, the video will contain ranges of correlated images, possibly using some images several times. The underlying idea is that the growth process of microtubules is assumed not to be linear.

## 4. EXPERIMENTAL RESULTS

We presents here our experimental results on ninety-eight microtubule images captured by an electronic microscope. Test images are obtained from different samples of microtubules. All of those images contain the plus-end of microtubules, which has the so-called α subunit exposed. Due to different imaging conditions and different growth stages of microtubules, plus-ends demonstrate rather big varitations in their appearances and locations across these images. Therefore, image registration was conducted to align these candidate images.

To simplify computation, manual pre-processing of images was conducted to translate the plus-end to center of an image. The illustration of this process is presented in Figure 2. One may find the size of images undergoing the processing shrinked. This is because the original microtubule images have different sizes. Our aim is to crop all images into the same size while have plus-end at the center of these images. The images were all circularly cropped to make sure the outer parts of the images did not interfere with the cross-correlation calculations in the last stage of the algorithm. Therefore, it became easy for our later stage processing of polar transformation on these images. The actual dimension of these cropped images are 1024*1024 pixels.

With the images normalized regarding plus-end alignment, the next step was to rotate candidate images in accordance with the pre-designated reference image. The reference image was also used as the initial frame for the output video. An efficient approach was devised to automatically rotate images. That is, we transformed all available images, except the reference image, into the polar form. Therefore, rotation in the cardesian coordinates translates to shifting in the polar coordinates. Figure 3 (a) and (b) demonstrate two original images before the coordinate transformation. The results after the polar transformation are shown in Figure 3 (c) and (d) respectively. The dimension of the polar version of the image was set to 720*200 pixels. 720 accounts for sampling the angles over a total of 360 degrees with 0.5 degrees steps. Furthermore, 200 steps are taken for the radius. The determination of an optimal rotation angle was made by comparisons of NCC values of the rotated versions of an image and the reference image. If the rotated images resembled the reference image more, their NCC values were closer to one. When the NCC values reached their maximum, we took the corresponding rotation angle as the optimal one.

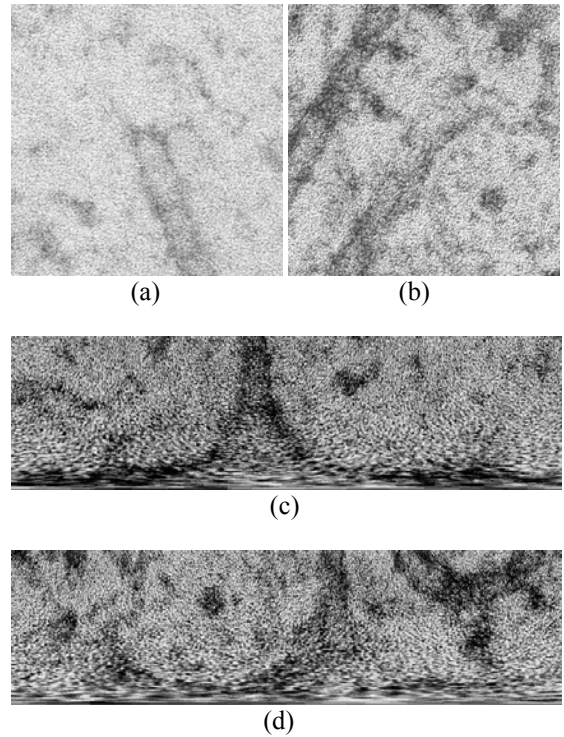

(a)          (b)

(c)

(d)

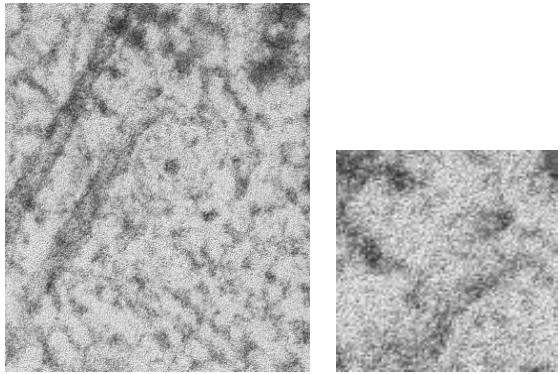

Figure 2 Illustration of the manual preprocessing of microtubule images. (a) Before processing, the plus-end locates at the top-right part of the image. (b) After processing, the plus-end is put at the center of the image.

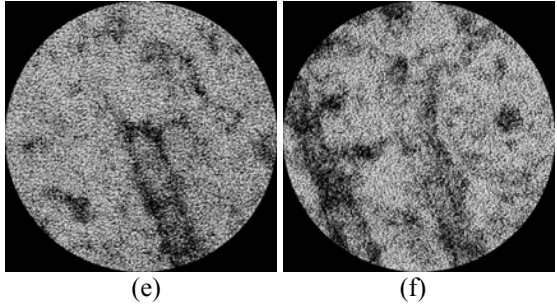

(e)                    (f)

Figure 3: The alignment of two microtubule images. a) and b) the two original images; c) and d) the polar versions of the normalised a) and b), of which d) will be shifted horizontally to find the highest NCC value; e) and f) the images after normalisation and rotation.

The final stage for our method was to generate a sequence of images that were pairwise highly correlated. For this purpose, we focused on the center part of the cropped image, and then calculated the normalized cross correlation of a candidate image and the reference image. The image which had the highest evaluation of NCC value was chosen to be the next key frame. Figure 4 demonstrates a sequence of key frames in our result video.

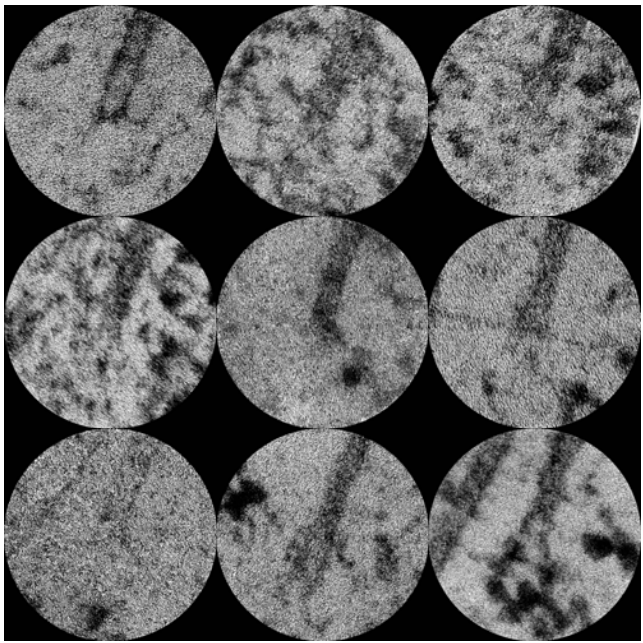

**Figure 4** Nine continuous key frames that are used for creation of the video.

## 5. CONCLUSIONS AND FUTURE WORK

In this work, we presented an efficient approach to explore the development process of microtubules. Our intention is to integrate static images taken at different stages to a video to simulate the process. Due to the confinement of electronic imaging and tissue preparation, the images obtained were highly noisy and randomly aligned. To get a consistent simulation of the growth of microtubule, we employed normalized cross correlation as the similarity measure to register images. For the ease of computation, we assumed that images could be mapped by a rotation transformation. By shifting the candidate image and sliding it through the reference image, the optimal angle can be obtained when the normalized cross correlation between the shifted version of the candidate image and the reference image arrives at its maximization. The aligned image can then be selected and combined into a proper video.

Although our preliminary method is effective and intuitive in implementation, there are still some improvements needed in future work, especially when we want to get a more computationally efficient and realistic simulation.

### 5.1 Allow More General Transformation Assumption

As we mentioned before, for the purpose of easy computation, only rotation of objects is considered in our work. This assumption is actually a good basis for the preliminary work. However, this assumption may not hold for all cased. In the following figure, it is clear that the plus-ends in the middle of the pictures have definitely different visual appearances. The right hand side one appears in the later stage of the growth process of a microtubule.

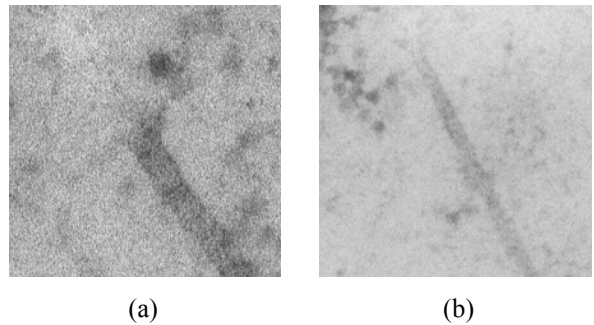

(a)                    (b)

**Figure 5** Comparison of two plus-ends (a) of smaller length and (b) with large length.

For more precise registration of these microtubule images, we need to consider some more general transformations, say, affine transformations, which includes

rotation, scale and translation. A general affine transformation can be formalized as the following:

$$\begin{pmatrix} x' \\ y' \\ 1 \end{pmatrix} = \begin{pmatrix} m_{11} & m_{12} & m_{13} \\ m_{21} & m_{22} & m_{23} \\ 0 & 0 & 1 \end{pmatrix} \begin{pmatrix} x \\ y \\ 1 \end{pmatrix} \quad (4)$$

where $(x, y)^T$ denotes the coordinates of the original point and $(x', y')^T$ denotes the coordinates of the transformed point. $m_{ij}$ denotes a transformation coefficient.

*5.2 Speed up the Template Matching*

As far as template matching is concerned, normalized cross correlation is often the adopted measure due to its robustness with respect to photometric variations. But with large image size or template size, the matching process can be computationally very expensive. How to speed up the basic algorithm and make an appropriate adaptation is the second critical issue, which needs to be solved in our future work.

A close investigation of the image data reviews that those images are significantly corrupted by noise. As shown in Figure 2, the object of interest, the plus-end, only occupies a small area of the image. It seems futile to compare the noise patterns by NCC due to its ignorance of structure information. A worthy trial is to de-noise these images, detect the dominant object in these images and then match these deformed objects across images. For the task of object matching, the method of deformable models seems to be a good choice.

*5.3 Devise an Appropriate Similarity Measure for Temporally Related Images*

In the literature of image registration, one fundamental problem is how to align two individual images. For most temple-matching problems, typical image registration methods work pretty well. The peculiarity of our problem, however, is a natural image sequence, which has an inherent temporal coherence. That characteristic was ignored in our previous work. Therefore, devising a special similarity measure for temporally related images becomes one of our major future works.

Suppose we have a sequence of temporally related images $\{I_1, I_2, I_3,...,I_n\}$. Given a fixed initial image $I_1$, the two critical properties the metric for measuring the self-similarity of the natural sequence should have include:

I. $P(I_n | I_{n-1}) \geq P(I_n | I_{n-2}) \geq ... \geq P(I_n | I_1)$. That is, the probability of the image $I_n$ generated by an intermediate image $I_i$ monotonically decreases with the deceasing of the subscript.

II. $P(I_n, I_{n-1},...I_2 | I_1)$, the probability of the whole image sequence with the initial image given, should have its maximal value for an optimal configuration of the sequence.

To model the temporal relationships inherent inside a natural image sequence, we assume that an image only dependent on its two adjacent neighbors. This assumption can be formalized as:

$$P(I_i | I_{i-1}, I_{i-2},...I_1) = P(I_i | I_{i-1}, I_{i-2})$$
$$= \frac{P(I_i, I_{i-1} | I_{i-2})P(I_{i-2})}{P(I_{i-1}, I_{i-2})} \quad (5)$$
$$= \frac{P(I_i, I_{i-1} | I_{i-2})}{P(I_{i-1} | I_{i-2})}$$

Figure 6 presents a graphic illustration of the above probabilistic dependency. One can see from the Eq.(5) that beside the probability of $P(I_i | I_{i-1})$ one similarity measure should also give evaluation of $P(I_i, I_{i-1} | I_i)$. Since $I_1$ is manually designated in the pre-processing stage, the generation probability of the whole sequence of images can be evaluated iteratively as:

$$P(I_n, I_{n-1},..., I_2 | I_1) = P(I_n | I_{n-1},..., I_2, I_1) P(I_{n-1}, I_{n-2},..., I_2 | I_1)$$
$$= P(I_n | I_{n-1}, I_{n-2}) P(I_{n-1}, I_{n-2},..., I_2 | I_1)$$
$$= \frac{P(I_i, I_{i-1} | I_{i-2})}{P(I_{i-1} | I_{i-2})} P(I_{n-1}, I_{n-2},..., I_2 | I_1)$$
(7)

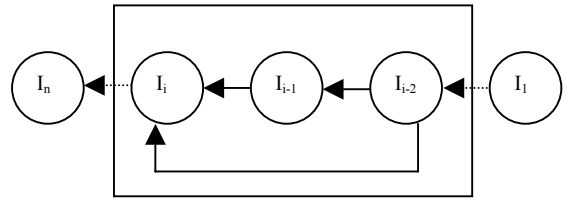

**Figure 6** Probabilistic dependencies between adjacent images.

Therefore, our task becomes how to determine an efficient template matching measure, which meets the two criteria, mentioned above.

*5.4 Improve the Performance of Exhaustive Searching*

Another fundamental problem in image registration is how to devise an efficient search strategy because of the large computational costs associated with many of the matching features and similarity measures. Generally speaking, the search space is the class of transformations from which we would like to find the optimal transformations to align the images. Due to our potential consideration of more general affine transformations, selection of the best search strategy becomes more critical for the success of our method. One promising approach is to limit the search to salient points [5].

Our aim becomes to find a sufficient condition that is capable of rapidly pruning those candidates that could not provide a better similarity measure with respect to the current best candidate. A promising approach can be obtained exploiting an upper bound of the similarity measure. This upper bound must be computed efficiently. We expect this approach can yield a significant reduction of operations compared to brute force evaluation of the similarity measure and allow for reducing the overall number of operations required to carry out exhaustive searches.